\documentclass{article}
\usepackage{spconf,amsmath,graphicx}
\usepackage{tabularx}

\newcommand{\dop}{\textit{domain-prompts}} 
\usepackage{makecell}
\usepackage{multirow}
\usepackage{xcolor}
\usepackage{booktabs}
\usepackage{amsmath}
\usepackage{array,etoolbox}
\preto\tabular{\setcounter{magicrownumbers}{0}}
\newcounter{magicrownumbers}
\def\rownumber{}

\title{Prompt Tuning GPT-2 language model for parameter-efficient \\ domain adaptation of ASR systems}
\name{Saket Dingliwal, Ashish Shenoy*\thanks{*work carried out while working at Amazon}, Sravan Bodapati, Ankur Gandhe, Ravi Teja Gadde, Katrin Kirchhoff}
\address{\{skdin, ashenoy, sravanb, aggandhe, gadderav, katrinki\}@amazon.com} 

\begin{document}
%
\maketitle

\begin{abstract}

Automatic Speech Recognition (ASR) systems have found their use in numerous industrial applications in very diverse domains creating a need to adapt to new domains with small memory and deployment overhead.
In this work, we introduce \dop{}, a methodology that involves training a small number of domain embedding parameters to prime a Transformer-based Language Model (LM) to a particular domain. 
Using this domain-adapted LM for rescoring ASR hypotheses can achieve 7-13\% WER reduction for a new domain with just 1000 unlabeled textual domain-specific sentences. This improvement is comparable or even better than fully fine-tuned models even though just $0.02\%$ of the parameters of the base LM are updated. 
Additionally, our method is deployment-friendly as the learnt domain embeddings are prefixed to the input to the model rather than changing the base model architecture. Therefore, our method is an ideal choice for on-the-fly adaptation of LMs used in ASR systems to progressively scale it to new domains. 

\end{abstract}


\section{Introduction}
\begin{figure*}
    \centering
    \includegraphics[width=0.72\linewidth]{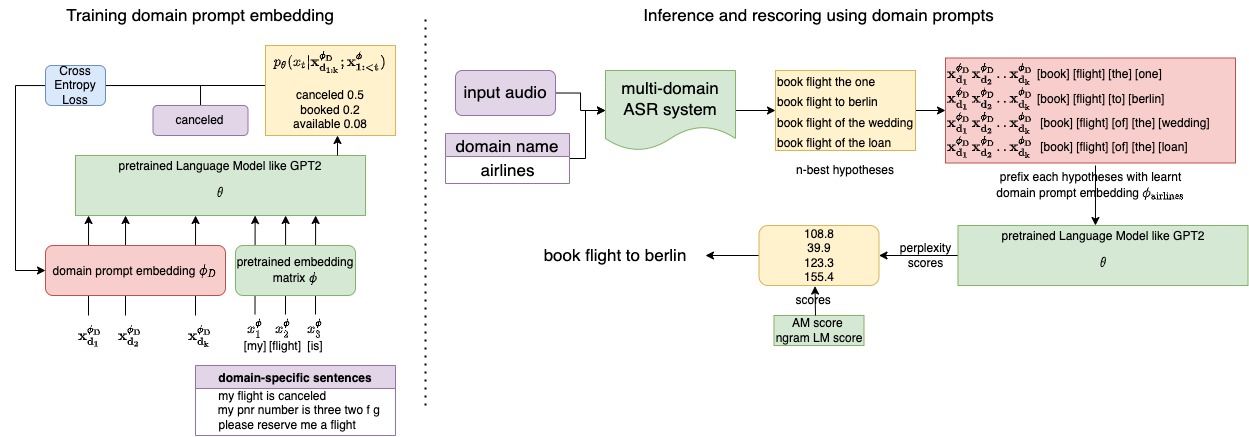}
    \caption{Domain Prompts: training (left) and inference (right) methodology for domain-adaptation}
    \label{fig:meth}
\end{figure*}

Automatic Speech Recognition (ASR) systems form a key component of various products across industry. With recent advancements \cite{graves2012sequence, chan2016listen, zhang2020Transformer}, they have been deployed in a wide range of domains, including healthcare, travel reservations, and customer services etc.  
A typical technique to improve the performance of these systems is to do a rescoring of the $n$-best hypotheses with an external Language Model (LM) \cite{chan2016listen}.
Recent pretrained Transformer-based LMs such as GPT-2 \cite{radford2019language} and BERT \cite{devlin2018bert}
have been shown \cite{irie2019language} to be more effective than conventional LSTM based LMs for rescoring. However, their use in an industrial ASR system that needs to incrementally support new domains poses the following challenge. As showcased in \cite{shenoy21_interspeech, Shenoy_2021}, domain-specific data is useful for improving performance in a domain. However, retraining or maintaining copies of Transformer-based LMs for each domain separately is not scalable as updating and storing millions of parameters comes with a large cost. Therefore, a need for an efficient domain-adaptation method for such LMs is evident. 
\cite{9746748, 9688109, jain2020contextual} used external knowledge, memory and context respectively to improve performance in specific difficult domains, while 
\cite{park2010improved, alumae2013multi} adapted the neural LM used within the system. However, to the best of our knowledge, ours is the first work to propose and study methods to do efficient domain-adaptation of Transformer-based LMs to benefit ASR systems. Language modeling literature \cite{pfeiffer2020adapterhub, lester2021power, li2021prefix} introduced novel methodologies to solve a related problem of efficiently adapting such LMs to specific tasks. Instead of fine-tuning and storing millions of parameters for each task, they propose augmenting the frozen task-agnostic model with a handful of task-specific trainable parameters. 
For example, AdapterHub \cite{pfeiffer2020adapterhub} introduced new task-specific layers in conjunction to frozen pre-trained weights of LMs. 
More recent models, such as GPT-3 \cite{brown2020language}, are able to solve new tasks with the help of just the textual descriptions of the task (called prompts). 

Extending \cite{DBLP:journals/corr/abs-2110-06502}, the focus of this work is to adapt such LMs to different domains of the same task rather than solving multiple tasks. Our objective is to learn a small set of domain-specific parameters to score ASR hypotheses better than the base Transformer-based LM without the domain data. Drawing ideas from prompt-tuning \cite{lester2021power} for task adaptation, we introduce \dop{} for our goal.
We define \dop{} as domain-specific embeddings, which when prefixed to the sequence of token embeddings, and passed through a pretrained Transformer LM, return the probability distribution of the next token, close to that given by a fully domain-adapted LM. 
Our main contributions are summarized as follows: 
(1) we introduce a new methodology \dop{}, which is the first attempt to apply prompt-tuning for parameter-efficient domain-adaptation of Transformer-based LMs for their use in ASR systems, 
(2) In new domains with limited data, we demonstrate that rescoring ASR hypotheses with LM adapted using our method can achieve 7-13\% WER reduction while using a handful of additional domain-specific parameters
(3) Along with saving memory and training cost, \dop{} can match or even beat the performance of fully fine-tuned models with no change to the deployment of the base model, thereby making it the ideal choice for on-the-fly domain adaptation for industrial ASR systems.


\section{methodology}
\label{sec:meth}
GPT-3 \cite{brown2020language} introduced natural-language \textit{prompts} as textual descriptions of a task. For example - prefixing "translate the sentence to French " to the input for the machine translation task. 
In prompt-tuning \cite{lester2021power}, rather than designing these prompts manually, the model learns their embeddings using a few labeled examples from the task.  
We demonstrate that such embeddings can also be learnt for different domains, i.e., we can prefix a sentence with additional domain-specific embedding vectors such that it improves the perplexity of the sentences from that domain. We use the self-supervised task of predicting the next token in unlabeled domain-specific text for training. 
An unlabelled sentence is a sequence of $T$ tokens $\{x_1, x_2 \dots x_{T}\}$. Let $\mathbf{x_{1:T}^{\phi}} $ be the corresponding concatenation of $d$-dimensional embedding vectors for these tokens, given by the embedding matrix parameterized by $\phi$. 
These vectors are propagated through multiple Transformer layers before taking a softmax to obtain the probability distribution over the vocabulary for the next possible token.
Mathematically, we denote the probability of predicting $x_t$ token at $t^{\text{th}}$ time-step  as  $p_{\theta}(x_t | \mathbf{x_{1:<t}^\phi})$ where $\theta$ represents all the parameters in the Transformer layers. Both $\theta$ and $\phi$ have large dimensions and are trained together on a large corpus of text.

In our method, for any domain $D$, we begin with pretrained $\{\theta, \phi\}$ and introduce a small number of additional parameters $\phi_D$ in the form of $k$ $d$-dimensional embedding vectors $[\mathbf{x_{d_{1}}^{\phi_D}}, \mathbf{x_{d_{2}}^{\phi_D}} \dots \mathbf{x_{d_{k}}^{\phi_D}}]$ concatenated together as $ \mathbf{x_{d_{1:k}}^{\phi_D}}$. We prefix them to each sentence while predicting the next token at each time step. While training, we keep $\{\theta, \phi\}$ fixed and learn $\phi_D$ by minimizing the cross-entropy loss between the true token value $x_t$ and its predicted probability. Eq. \ref{eq:loss} represents the loss for one such sequence and we add the loss value for all the sentences from the domain. 
\begin{equation}\label{eq:loss}
\phi_D =   \operatorname*{argmin}\limits_{\phi_D} \sum\limits_{t=1}^{t=T} - \log p_{\theta}(x_t | \mathbf{x_{d_{1:k}}^{\phi_D}};\mathbf{x_{1:<t}^{\phi}}) 
\end{equation}

We hypothesize that the self-attention layer in Transformer-based LMs will create an interaction between $\phi_{D}$ and the embeddings of the tokens from sentences, thereby improving the score to cater to the domain $D$. 
During inference, 
we prefix the trained \dop{} ($\mathbf{x_{d_{1:k}}^{\phi_D}}$) of the corresponding domain to the hypotheses from the ASR system and use perplexity scores from the model for rescoring as explained in Fig.~\ref{fig:meth}.
Further, in our implementation, we used gradient descent to minimize the loss and instead of initializing the parameters $\phi_{D}$ randomly, we begin with token embeddings (using $\phi$ from pretrained model) of the $k$ most frequent words in the training sentences of the domain, following prior work \cite{lester2021power}. Also, to ensure inference latency does not increase due to additional computations, we use caching to store the state of the Transformer after propagating the $k$ \dop{}  through the Transformer layers. These domain-embeddings are constant for all the hypotheses in the domain and hence its forward pass through Transformer layers can be precomputed to ensure the latency for scoring a hypothesis is same for both base and adapted versions of the Transformer LM. 

\vspace{-0.3cm}
\section{Experimental Setup}
\label{sec:exp}
\vspace{-0.2cm}

To test the effectiveness of the proposed \dop{}, we run extensive experimentation with numerous adaptation baselines with four different domains, model sizes, initializations and training set sizes. We use two versions of GPT-2 architecture \cite{radford2019language} as our base models: (1) \textit{gpt2} (117M (million) parameters) (2) \textit{gpt2-medium} (345M parameters). We conduct all our experiments on an AWS ec2 instance with 8 Tesla V100 GPUs using a hybrid ASR system. Such a system consists of an Acoustic Model (AM), and two different LMs. \textit{First pass LM} is an ngram LM which is directly composed with lattice from the AM while \textit{second pass LM} is a Neural LM and is used for rescoring the $n$-best hypotheses decoded from the lattice. 
Our AM is trained on 12k hours of audio. Our first pass LM is a 4-gram model with Kneser-Ney (KN) \cite{kneser1995improved} smoothing with vocabulary size of 500k words trained on a heterogeneous corpus of text. We use scores from different second pass LMs (and interpolate with scores from the AM and the first pass LM) to rescore $n$-best lists with $n=10$. Our performance metric is Word Error Rate (WER) and we report WER Reduction \% (WERR) over the baseline. 
We evaluate under two different settings representative of any industrial multi-domain ASR system: (1) \textit{New domains with limited data}: To simulate the scenario of on-the-fly adaptation to new/unseen domains, we use only 1k domain-specific sentences to adapt the second pass LM. (2) \textit{Domains with large data}: 50k domain-specific sentences are added to the training set of both the first and the second pass LM.




\noindent\textbf{Dataset}:  For our experiments, we need domain-wise (i) textual data for LM adaptation, (ii) audio data for evaluation.
We were not able to find any public dataset that can be split domain-wise and meets both the criteria.
Therefore, we use in-house datasets from four different domains with all Personal Identifiable Information (PII) removed. For textual data, we use 1k and 50k domain-specific conversational sentences in the two settings respectively. This data is split into 80:20 as train and dev set respectively. The perplexity on the dev set is used for tuning the hyperparameters like learning rate etc for all the baselines. For evaluation, we use 500 labelled 8khz audios per domain which are single utterances from a conversational task-oriented dialog system. 


\noindent\textbf{Baselines}: The different choices of second pass LM used for comparison in Table \ref{tab:results} are defined below: 
(1) \textit{no rescoring} (baseline): Use the 1-best hypothesis with no second pass LM.
(2) \textit{LSTM LM}: 2-layer LSTM based LM with embedding dimension ($d$), hidden dimension ($h$) and a word-based tokenizer with vocabulary size ($V$). 
(3) \textit{no adaptation}: Out-of-the-box Transformer-based LMs without the use of any domain data or any parameter update. 
(4) \textit{tuning embedding layer}: Update the parameters of the embedding matrix ($\phi$) using domain-specific data while keeping $\theta$ fixed.
(5) \textit{prompt-designing}: Prefix manually-defined prompts to the hypotheses without any training. 20 most frequent words from the domain were used as fixed prompts.
(6) \textit{domain-embedding}  \cite{shenoy21_interspeech}: Learn the embedding of a special domain token. It is equivalent to $k=1$ in our method.  
(7) \textit{domain Adapters}: Adapters \cite{pfeiffer2020adapterhub} are typically used for parameter-efficient task adaptation of Transformer-based LM. However, we train \textit{domain Adapters} here using the self-supervised task of next token prediction with different reduction factors ($c$).
(8) \dop{}: Update the domain embeddings ($\phi_D$) for different values of $k$ and initializations of $\phi_D$. 
(9) \textit{full fine-tuning}: Update all the parameters ($\theta, \phi$) of the base LM.
(10) \textit{oracle}: pick hypothesis in the $n$-best list with minimum WER to know the upper bound of improvements through rescoring.



\vspace{-0.3cm}
\section{Results and Discussion}
\label{sec:disc}
\vspace{-0.2cm}

\begin{table}[]
\centering
\caption{Generating text from \textit{gpt2} adapted to airlines domain}
\resizebox{\linewidth}{!}{
\begin{tabular}{|l|l|}
\hline
Input tokens                   & "hello how are you"                                                                                                                                                              \\ \hline
no-adaptation    & \begin{tabular}[c]{@{}l@{}}hello how are you doing?\\ I'm really happy with the results.  \end{tabular}             \\ \hline
full-fine-tuning  & \begin{tabular}[c]{@{}l@{}}hello how are you able to get a new flight \\ I'm flying from London Heathrow to Dubai \end{tabular} \\ \hline
domain-prompts & \begin{tabular}[c]{@{}l@{}}hello how are you able to get a refund on the flight\\ I'm flying from Glasgow to Madrid today\end{tabular}                    \\ \hline
\end{tabular}
}
\label{tab:gentext}
\end{table}

\begin{table*}
\centering
\caption{Comparison of different domain-adaptation methods for different domains in parameter count and WERR\% metric.}
\label{tab:results}
\resizebox{0.97\linewidth}{!}{
\begin{tabular}{|@{\makebox[3em][r]{\rownumber\space}}|l|c|c||c|c|c|c||c|c|c|c|} 
\hline
\begin{tabular}[c]{@{}l@{}}\\\end{tabular}                                    &                     &       \textbf{\# additional }      & \multicolumn{4}{c||}{\textbf{Low data setting} (1k sentences)}                    & \multicolumn{4}{c|}{\textbf{Large data setting} (50k sentences)}                  \\ 
\cline{1-2} \cline{4-11}

\textbf{domain adaptation methods} & \textbf{base model} &  \textbf{domain-specific} & \multicolumn{4}{c||}{\textbf{WER Relative \% } $\uparrow$}                                   & \multicolumn{4}{c|}{\textbf{WER Relative \% } $\uparrow$}                                    \\ 
\cline{1-2} \cline{4-11}
                                                                              &                     &   \textbf{params}              & \textbf{airlines} & \textbf{fastfood} & \textbf{healthcare} & \textbf{insurance} & \textbf{airlines} & \textbf{fastfood} & \textbf{healthcare} & \textbf{insurance}  \\ 
\hline
 \gdef\rownumber{\stepcounter{magicrownumbers}\arabic{magicrownumbers}} 
no rescoring                                                                  & -                   & 0                  & -                 & -                 & -                   & -                  & -                 & -                 & -                   & -                   \\ 
\Xhline{4\arrayrulewidth}
LSTM LM ($d=h=256, V=15k$)                                                 & LSTM                & 8.7M               & 0                 & 0                 & 0                   & 0                  & 4.8               & 0.8               & 0                   & 6.2                 \\ 
\hline
LSTM LM ($d=h=512, V=229k$)                                                 & wiki103-LSTM                &        121M            &       3.1            &          0         &     0                &      3.9              &        7.3           &         1.7          &        0             &     6.2                \\ 
\Xhline{4\arrayrulewidth}
no adaptation~                                                                & gpt2                & 0                  & 0.5               & 2.0               & 2.4                 & 2.7                & 2.4               & 0.8               & 0.8                 & 3.3                 \\ 
\hline
no adaptation~                                                                & gpt2-medium         & 0                  & 3.6               & 3.4               & 3.4                 & 4.8                & 4.9               & 3.4               & 0.8                 & 8.2                 \\ 
\hline
no adaptation~                                                                & dialog-gpt2-medium  & 0                  & 3.1               & 1.5               & 1.0                 & 3.2                & 4.9               & 2.6               & 0                   & 5.1                 \\ 
\Xhline{4\arrayrulewidth}
tuning embedding layer~                                                       & gpt2                & 40M            &  3.6                 &       4.9            &        3.9             &  6.0                  &       7.3            &        9.4           &  0                    &         6.2            \\  
\Xhline{4\arrayrulewidth}
prompt-designing~                                                           & gpt2                &     0        &              0.5     &       0          &        2.4             &          2.7          &     3.7             &          1.7         &  0                    &         4.1            \\  
\Xhline{4\arrayrulewidth}
domain-embedding~                                                       & gpt2                &  768            &             0.5      &        2.0           &         3.4            &        4.3            &        3.7          &      1.7             &           0           &       3.1             \\  
\Xhline{4\arrayrulewidth}
domain Adapter ($c=512$)~                                                       & gpt2                & 0.3M               & 4.6               & 5.9               & 3.4                 & 5.9                & 6.1               & 7.7               & 0.8                 & 6.2                 \\ 
\hline
domain Adapter ($c=16$)~                                                        & gpt2                & 1.1M               & 4.6               & 7.8               & 5.8                 & 6.4                & 8.5               & 11.1              & 0.8                 & 8.2                 \\ 
\hline
domain Adapter ($c=512$)~                                                       & gpt2-medium         & 1M                 & 7.1               & 10.7              & 6.3                 & 8.1                & 9.8               & 8.6               & 1.6                 & 10.3                \\ 
\hline
domain Adapter ($c=16$)~                                                        & gpt2-medium         & 3.6M               & 7.6               &        9.8           &       4.8              &   7.0                  & 8.5               & 12.0              & 3.2                 & 10.3                \\ 
\Xhline{4\arrayrulewidth}
domain-prompts ($k=10$, vocab init)~~                                           & gpt2                & 7680    & 3.8               & 7.3               & 5.3                 & 3.2                & 6.1               & 7.7               & 3.2                 & 7.2                 \\ 
\hline
domain-prompts ($k=50$, random init)                                            & gpt2                & 0.04M     
 &      6.1             &      8.8             &      6.3               &         6.5           &       6.1            &       8.6            &       2.4              &             8.2        \\ 
\hline
domain-prompts ($k=50$, vocab init)                                             & gpt2                & 0.04M         
& 5.1               & 8.8               & 6.8                 & 7.0                & 8.5               & 9.4               & 4.0                 & 9.2                 \\ 
\hline
domain-prompts ($k=200$, vocab init)                                            & gpt2                & 0.16M              & 6.1               & 9.3               & 7.3                 & 5.9                & 8.5               & 9.2               & 4.0                 & 10.3                \\ 
\hline
domain-prompts ($k=50$, vocab init)                                             & gpt2-medium         & 0.05M              & \textbf{8.1}      & \textbf{13.1}     & \textbf{7.7}        & \textbf{8.1}       & \textbf{11.0}     & 11.1              & 5.7                 & \textbf{12.4}       \\ 
\Xhline{4\arrayrulewidth}
full-fine-tuning                                                              & gpt2                & 117M               & 6.6               & 9.8               & 6.8                 & 6.4                & 8.5               & 12.9              & 4.8                 & 12.3                \\ 
\hline
full-fine-tuning                                                              & gpt2-medium         & 345M               & 7.1               & 11.2              & 7.2                 & 7.0                & 8.5               & \textbf{16.2}     & \textbf{7.2}        & \textbf{12.4 }               \\
\Xhline{4\arrayrulewidth}
oracle                                    & - & -                          &  22.0          & 32.8               & 29.8               & 21.1          & 39.3               & 38.6     & 36.2        &  34.7               \\
\Xhline{4\arrayrulewidth}
\end{tabular}
}
\end{table*}

Domain adaptation methods are used to prime the LM to a particular domain. As shown in Table \ref{tab:gentext}, the LM adapted using \dop{} learnt from airlines data, completes sentence to a very domain-specific utterance in contrast to the out-of-the-box LM extending the input to a generic sentence. We showcase some qualitative results in Table \ref{tab:qual}, where perplexity scores from vanilla \textit{gpt2} and \dop{} ($k=50$) for similar sounding hypotheses is provided. These examples indicate how domain information helps to disambiguate to choose the right hypothesis.
We summarize the WERR scores of all our methods in the two different settings in Table \ref{tab:results}. 
The first column contains the name of the methodology and its hyper-parameter, second column is the base model used while the third column represents the number of domain-specific trainable parameters needed in addition to the base model. 
Note that the goal of the experiments is not only to find the best performing method, but also to discover settings that achieve optimal performance with the minimal number of additional parameters. This is a critical decision point for systems to scale to potentially hundreds of domains as storage and training cost are directly linked to the number of domain-specific parameters.

The adaptation to domain-specific data is useful for performance in all the domains in both the settings (row 4 vs. 19 or row 3 vs. 18). Even rescoring with \textit{dialog-gpt2-medium} \cite{zhang2019dialogpt} (row 5), which is pretrained on a large dialog corpus is not as effective as adaptation to small amounts of domain data. The WERR numbers vary across different domains but the relative performance for different domain-adaptation methods is consistent across all domains. The domains are fairly different from each other. Domains like healthcare have a large number of unseen technical words and hence improvements through rescoring are relatively small.  In the large-data-setting, the domain-specific data is also added to first pass LM and hence the quality of the $n$-best hypotheses is better, which leads to larger performance improvements through rescoring (row 20). 
Similar to results in \cite{irie2019language}, we observe Transformer-based LMs perform better than LSTM based methods (row 1 and 2 vs. 18). For fair comparison, we increase the size of the LSTM model  and pretrain it with wiki-text-103 \cite{merity2016pointer} (row 2), but it still cannot match Transformer models. Also, as we increase the size of the Transformer, the performance improves (row 18 vs. 19), further indicating the need for parameter-efficient adaptation methods. Our main conclusions about our method are as follows:

\noindent\textbf{Domain prompts are the most parameter efficient}: \dop{} uses $<0.02$\% of parameters of the base model per domain to achieve performance comparable to domain-specific fully fine-tuned models with millions of parameters (row 17 vs. 19). Although fine-tuned models perform better than our method in the large-data-setting, their improvement comes at the cost of deploying separate models for each domain. This is expected when adequate data is available, large number of domain-specific parameters can capture larger amount of domain information.  Adapters which are commonly used for their efficiency, have limited efficacy when compared to our method. \textit{Domain prompts} with 20 times less parameters, can beat its performance (row 11 vs. 17). Further, Adapters have another limitation that their number of parameters scale with the number of layers in the base Transformer model (\textit{gpt2} vs.\textit{gpt2-medium}) while \dop{} depends only on the embedding dimension $d$.
Fine-tuning a subset of parameters in Transformer-based LM (row 6) is not effective as manually selecting a subset of most influential parameters is difficult and performs worse than our method in both performance and cost. 
Methods like fixing prompts or training a single domain embedding vector use no or very small number of parameters but their improvements are only marginal over unadapted base LM (row 3 vs. 7 and 8). 



\noindent\textbf{Domain prompts achieves best performance for new/unseen domains}:  This setting represents common practical applications where a new domain with limited amount of available data needs to be added to the ASR system. Here, \dop{} can reap both the benefits: (1) rich pretraining of Transformer based LM (2) no overfitting on limited number of examples. This is evident from fact that fine-tuned \textit{gpt2} performs slightly better than corresponding prompt-tuned version (row 16 vs. 18) while opposite is true for \textit{gpt2-medium} (row 17 vs. 19) indicating updating large number of parameters is prone to overfitting. Hence, \dop{} presents an ideal case to capture all the necessary domain specific information from 1k examples in its limited domain-specific parameters and achieve 7-13\% WERR improvement. 

\noindent\textbf{Domain prompts are deployment friendly}: In addition to performance and cost benefits, \dop{} can easily be used for new domains without having to deploy new models or introducing new architectures. \textit{Domain prompts} are prefixed to the input keeping all the base model parameters unchanged, while all other adaptation methods require updating the parameters inside the base model architecture. 

\noindent\textbf{Domain prompts provides hyper-parameter ($k$) to trade-off performance and cost}: Comparing row 13, 14 and 16 in Table \ref{tab:results}, we see that the performance of models improves as we increase $k$, although the improvements saturate. This provides ASR system developers a parameter to control cost as per their requirements and availability of domain data. 

\vspace{0.1cm}
\noindent\textbf{Initialization with common vocabulary token embeddings helps}: Comparing rows 14 and 15, initializing $\phi_D$ with token embedding of most frequent domain words gives marginal improvements. Since these words are representative of the domain, they prove to be a useful starting point.

\begin{table}
\centering
\caption{Qualitative examples: domain-adapted \textit{gpt2} prefers hypothesis (green) over hypothesis with incorrect tokens (red)}
\label{tab:qual}
\resizebox{0.9\linewidth}{!}{
\begin{tabular}{|l|c|c|} 
\hline
\multicolumn{1}{|c|}{\textbf{hypothesis}} & \multicolumn{2}{c|}{\textbf{perplexity} $\downarrow$} \\
\hline
 & vanilla \textit{gpt2}  & adapted \textit{gpt2}~ \\ 
\hline

\multicolumn{1}{|l|}{\textbf{insurance}}                                                                        &   &                                                                                                                                                                                                      \\ 
\hline
\begin{tabular}[c]{@{}l@{}}i would like to retrieve my  \textcolor{red}{code}\\ i would like to retrieve my \textcolor[rgb]{0,0.502,0}{quote}\end{tabular}                                                                & \begin{tabular}[c]{@{}l@{}}\textbf{172.5}\\238.1\end{tabular} & \begin{tabular}[c]{@{}l@{}}40.4\\\textbf{21.4}\end{tabular}  \\
\hline

\multicolumn{1}{|l|}{\textbf{airlines}}                                                                                                                                                               &                                         &                               \\ 
\hline
\begin{tabular}[c]{@{}l@{}}what's the point \textcolor{red}{to tell you} for frequent flyer number \\what's the points \textcolor[rgb]{0,0.502,0}{tally} for frequent flyer number\end{tabular} & \begin{tabular}[c]{@{}l@{}}\textbf{313.5}\\598.9\end{tabular} & \begin{tabular}[c]{@{}l@{}}19.9\\\textbf{17.6}\end{tabular}  \\ 
\hline
\end{tabular}
}
\end{table}

\vspace{-0.3cm}

\section{Conclusion}
\vspace{-0.2cm}
\textit{Domain prompts} provides a scalable and parameter-efficient method to add domain information to Transformer based LMs. It saves storage and training cost without compromising performance. It also achieves the best performance for new domains with only handful of available examples. Rather than updating the base model parameters, the new parameters are added as prefixes to input, hence our method doesn't require model deployments per domain. Therefore, our method becomes an ideal choice for on-the-fly adaptation of second pass LMs for incrementally scaling the industrial ASR system to new domains with negligible overhead.

\vfill\pagebreak

\bibliographystyle{IEEEbib}
\bibliography{refs,anthology}

\end{document}